\documentclass[,12pt]{elsarticle}

\usepackage{amssymb}
\usepackage{textcomp}
\usepackage{siunitx}

\usepackage{booktabs}

\journal{Multimedia Tools and Applications}

\begin{document}
\begin{frontmatter}



\title{Mask-guided sample selection for Semi-Supervised Instance Segmentation}


\author{Miriam Bellver$^1$, Amaia Salvador$^2$, Jordi Torres$^1$ and Xavier Giro-i-Nieto$^2$
$^1$Barcelona Supercomputing Center (BSC) \\
Jordi Girona street, 29, 31, 08034 Barcelona, Spain \\
$^2$Universitat Polit\`ecnica de Catalunya (UPC)\\ 
Jordi Girona street, 1-3, 08034 Barcelona, Spain \\
\small{{\{miriam.bellver, jordi.torres\}@bsc.es}, \{amaia.salvador, xavier.giro\}@upc.edu} }
\address{}
\begin{abstract}

Image segmentation methods are usually trained with pixel-level annotations, which require significant human effort to collect. The most common solution to address this constraint is to implement weakly-supervised pipelines trained with lower forms of supervision, such as bounding boxes or scribbles. Another option are semi-supervised methods, which leverage a large amount of unlabeled data and a limited number of strongly-labeled samples. In this second setup, samples to be strongly-annotated can be selected randomly or with an active learning mechanism that chooses the ones that will maximize the model performance.
In this work, we propose a sample selection approach to decide which samples to annotate for semi-supervised instance segmentation. Our method consists in first predicting pseudo-masks for the unlabeled pool of samples, together with a score predicting the quality of the mask. This score is an estimate of the Intersection Over Union (IoU) of the segment with the ground truth mask. We study which samples are better to annotate given the quality score, and show how our approach outperforms a random selection, leading to improved performance for semi-supervised instance segmentation with low annotation budgets. 
\end{abstract}




\end{frontmatter}



\section{Introduction}
\label{sec:intro}

Instance segmentation is a popular task in computer vision in which a mask and a class category are predicted for each target object in a given image. Typically, high-performing models rely on large datasets of annotated data, which are expensive to obtain. This work extends our previous study that presented a semi-supervised scheme, which we refer as BASIS~(from \textbf{B}udget \textbf{A}ware \textbf{S}emi-supervised semantic and \textbf{I}nstance \textbf{S}egmentation)~\cite{bellver2019budget}. Given a low annotation budget, BASIS is able to outperform previous works on weakly or semi-supervised semantic and instance segmentation.

Figure \ref{fig:scheme} illustrates BASIS semi-supervised pipeline. It consists in training an \textit{annotation} network using only a few strongly-annotated samples, which is subsequently used to pseudo-annotate unlabeled or weakly-labeled samples. A second \textit{segmentation} network is trained with both the few strongly-annotated samples and the pseudo-annotations obtained by the \textit{annotation} network. In our previous work, the subset of samples to strongly-annotate was chosen randomly. In this work, we propose an alternative selection scheme, related to active learning methods, which leads to improved performance.

Our proposed method for sample selection consists in first training a model with a random subset of very few strongly-annotated images, and use it to obtain pseudo-annotations for a pool of unlabeled samples. Together with the pseudo-annotation, the model predicts a confidence score reflecting the quality of the masks produced by the network. This score informs us of how easy or difficult it is to segment the different object instances with the current model. Based on this score, a subset of samples from the pool is selected to be strongly-annotated by a human. Finally, as in BASIS, the \textit{annotation} network is trained with the available strongly-annotated samples. 

Our main contribution is the definition of a novel way to estimate the mask confidence score. Specifically, our model is trained to predict an estimation of the Intersection Over Union (IoU) of the pseudo-annotations produced with the corresponding ground truth masks. 
Given the predicted confidence score, we select which samples to strongly-annotate next given a limited annotation budget.  While IoU prediction has been used in previous works on object detection for filtering object proposals~\cite{jiang2018acquisition}, to the best of the authors' knowledge this work is the first one to use it as an active learning criterion. Moreover, we study the properties of the selected images, and conclude that the best images to annotate are those that are neither the easiest nor the most complicated of our dataset. With the Mask-guided sample selection strategy we reach higher performance compared to our BASIS baseline. 

The remainder of this manuscript is structured as follows: Section~\ref{sec:relatedwork} covers the previous works on weakly and semi-supervised instance segmentation, active learning pipelines and IoU prediction. Following, the benchmark of budget-aware segmentation is explained in Section~\ref{sec:benchmark}. Next, the semi-supervised architecture that we extend is reviewed in Section~\ref{sec:basis}. In Section~\ref{sec:iou-quality}, the IoU quality prediction pipeline is described. In Section~\ref{sec:experiments} the experimental validation is presented. Finally, Section~\ref{sec:conclusion} draws the conclusions of this work.

\section{Related Work}
\label{sec:relatedwork}

\noindent\textbf{Weakly-Supervised Instance Segmentation.} Few works have addressed weakly-supervised instance segmentation in computer vision. Bounding box labels have been exploited by \cite{khoreva2017simple}\cite{zhao2018pseudo}\cite{li2018weakly} to recursively generate and refine pseudo-labels from a weak-labeled set. These methods typically rely on bottom-up segment proposals~\cite{pont2017multiscale}\cite{rother2004grabcut}. In contrast with this approach, \cite{remez2018learning} proposes an adversarial scheme that learns to segment without using any object proposal technique. Although these works tackle weakly-supervised instance segmentation, their weak supervision consists in using bounding boxes, thus their main challenge resides in how to separate the foreground from the background within a bounding box. The first work that uses image-level supervision for weakly-supervised instance segmentation ~\cite{zhou2018weakly} detects peaks of Class Activation Maps~(CAMs)~\cite{zhou2016learning}, producing what they identify as Peak Response Maps~(PRMs). With them they generate a query to retrieve the best candidate among a set of pre-computed object proposals (MCG)~\cite{pont2017multiscale}. Recently, \cite{laradji2019masks} builds on PRMs by using the pseudo-masks to train Mask R-CNN~\cite{he2017mask} in a fully-supervised way, reaching better performance.

\noindent\textbf{Semi-Supervised Segmentation.} 
To the authors knowledge, only \cite{hu2017learning}\cite{li2018weakly} have tackled semi-supervised instance segmentation in still images. However, they assume a huge amount of weakly-labeled samples~(using MSCOCO~\cite{lin2014microsoft}). In our previous work, we focused on low-budget scenarios presenting the first results for semi-supervised instance segmentation for the Pascal VOC benchmark~\cite{everingham2010pascal} with no extra images from other datasets~\cite{bellver2019budget}. In this present work, we work with the same pipeline but extend it to a better selection of samples to be strongly-annotated for the semi-supervised scheme.

\noindent \textbf{Active learning}~\cite{settles2009active} consists in recursively selecting which samples to annotate in order to train a network. The goal of this approach is the reduction of the annotation cost, by only annotating those samples that will have more impact to the learning of the model. This acquires special relevance in contexts where annotating samples is very expensive, e.g., in image segmentation problems. Common active learning methods select samples according to two main criteria: how \textit{uncertain} and \textit{representative} a sample is. The \textit{uncertainty} is related to how informative a sample is with respect to the learning process. 

There are several methods that estimate the \textit{uncertainty}, e.g., dropout has been used to sample from the approximate posterior of a Bayesian CNN to calculate the uncertainty of predictions when varying the model~\cite{gal2016dropout}. This quantified metric can be used to request the annotation of subsequent training batches of data~\cite{gal2017deep}\cite{gorriz2017cost}. More recent methods have also used Bayesian CNNs to calculate the informativeness of images generated by a Generative Adversarial Network~(GAN)~\cite{mahapatra2018efficient} in order to add these samples to the training set.
Another method~\cite{efron1994introduction} is based on bootstrapping, and consists in training several networks with different subsets and calculate the variance in predictions across the different networks in order to estimate uncertainty~\cite{yang2017suggestive}. 

Some of the aforementioned methods not only base their selection on the \textit{uncertainty} criterion, but also on the \textit{representativeness} of a sample. This criterion is relevant to promote diversity among samples and to avoid redundancy. One strategy used in computer vision is to extract image descriptors with a CNN, and compare images with a cosine similarity metric~\cite{yang2017suggestive} to avoid picking very similar samples. Maximizing set coverage has also been studied~\cite{feige1998threshold}. Other metrics, such as \textit{content distance} have been used to quantify the distance of images based on their content to maximize content information~\cite{ozdemir2018learn}\cite{ozdemir2018active}.

Most of the above methods focus on image recognition and region labeling. The first works that handled active learning for large scale object detection~\cite{vijayanarasimhan2014large} used as active learning criterion the \textit{simple margin} selection method for SVMs~\cite{tong2001support}, which seeks points that most reduce the version space. More recently, methods rely on modern object detectors~\cite{redmon2016you}\cite{liu2016ssd}, but still are based on uncertainty indicators like least confidence or 1-vs-2~\cite{brust2018active}\cite{roy2018deep}. Notice that object detection is very close to the instance segmentation task addressed in this work. However, our sample selection criterion is based on the estimated quality of the different masks predicted for each image, instead of using classification scores as the previous approaches. 


\noindent \textbf{IoU prediction} IoU prediction has been used in recent works for filtering object proposals in object detection tasks~\cite{jiang2018acquisition}. Precisely, in \cite{jiang2018acquisition} the IoU between predicted bounding boxes and ground truth bounding boxes is estimated, and the authors argue that this score, in comparison to a class confidence score, considers the localization accuracy. In their work they show how their approach leads to improved performance. Similarly to this work, \cite{huang2019mask} predict the IoU between the predicted masks and the ground truth masks, and use this score to better filter object proposals for instance segmentation. 
In this direction, we propose to also predict the Intersection Over Union of the predicted masks with respect to the ground truth as a measure of the confidence of the prediction. 
\section{Benchmark for budget-aware segmentation}
\label{sec:benchmark}

As in our previous work~\cite{bellver2019budget}, we propose a unified analysis across different supervision setups and supervision signals for instance segmentation. Our motivation raises from the ultimate goal of weakly and semi-supervised techniques: the reduction of the annotation burden. 
We adopt the analysis framework from \cite{bearman2016s} and extend it to any supervision setup to compare to other works considering the total annotation cost.

We estimate the annotation cost of an image from a well-known dataset for semantic and instance segmentation: the  Pascal VOC dataset~\cite{everingham2010pascal}. Our study considers four level of supervision: image-level, image-level labels + object counts, bounding boxes, and full supervision (i.e.~pixel-wise masks).
The estimated costs are inferred from three statistical figures about the Pascal VOC dataset drawn from~\cite{bearman2016s}: a) on average 1.5 class categories are present in each image, b) on average there are 2.8 objects per image, and c) there is a total of 20 class categories. 
Hence, the budgets needed for each level of supervision are:

\noindent\textbf{Image-Level (IL):} According to \cite{bearman2016s}, the time to verify the presence of a class in an image is of 1 second. The annotation cost per image is determined by the total number of possible class categories (20 in Pascal VOC). Then, the cost is of $ t_{IL}\, =\,20\,\mathrm{classes/image}\,\times\,1 \mathrm{s/class}\,=\,20\,\mathrm{s/image}$.


\noindent\textbf{Image-Level + Counts (IL+C):} IL annotations can be enriched by the amount of instances of each object class.
This scheme was proposed in for weakly-supervised object localization~\cite{gao2017c}, in which they estimate that the counting increases the annotation time to 1.48s per class.

\noindent Hence, the time to annotate an image with image labels and counts is $t_{IL+C} =\, t_{IL}\,+\,1.5\,\mathrm{classes/image}\,\times\,1.48 \, \mathrm{s/class}\,=\,22.22\,\mathrm{s/image}$.

\noindent\textbf{Full supervision (Full):} We consider the annotation time reported in 
~\cite{bearman2016s} for instance segmentation: 
$t_{Full} = 18.5 \, \mathrm{classes/image}\times 1 \mathrm{s/class} \, + \, 2.8\,\mathrm{mask/image} \, \times \, 79 \, \mathrm{s/mask} \, = \, 239.7 \, \mathrm{s/image}$.

\noindent\textbf{Bounding Boxes (BB):} Recent techniques have cut the cost of annotating a bounding box to 7.0 s/box by clicking the most extreme points of the objects~\cite{papadopoulos2017extreme}. Following the same reasoning as for dense predictions, the cost of annotating a Pascal VOC image with bounding boxes is $t_{bb} = 18.5 \, \mathrm{classes/image}\times 1 \mathrm{s/class} \, + \,2.8\,\mathrm{bb/image} \, \times\, 7 \, \mathrm{s/bb} \, = \, 38.1 \, \mathrm{s/image}$.

Table~\ref{tab:summary_costs} summarizes the average cost of the different supervision signals for a single Pascal VOC image. In this work these annotation costs will be used as reference to compare between different configurations or to other works.

\begin{table}
\centering
\begin{tabular}{ccccc}
\toprule
    & \textbf{IL}  & \textbf{IL+C} & \textbf{Full}  & \textbf{BB}   \\
\bottomrule
Cost (s/image) & 20 & 22.22 &239.7  &38.1\\ 
\hline

\end{tabular}
\caption{Average annotation cost per image when using different types of supervision.}
\label{tab:summary_costs}
\end{table}

\section{BASIS}
\label{sec:basis}

\begin{figure}
  \centering
  \includegraphics[width=\columnwidth]{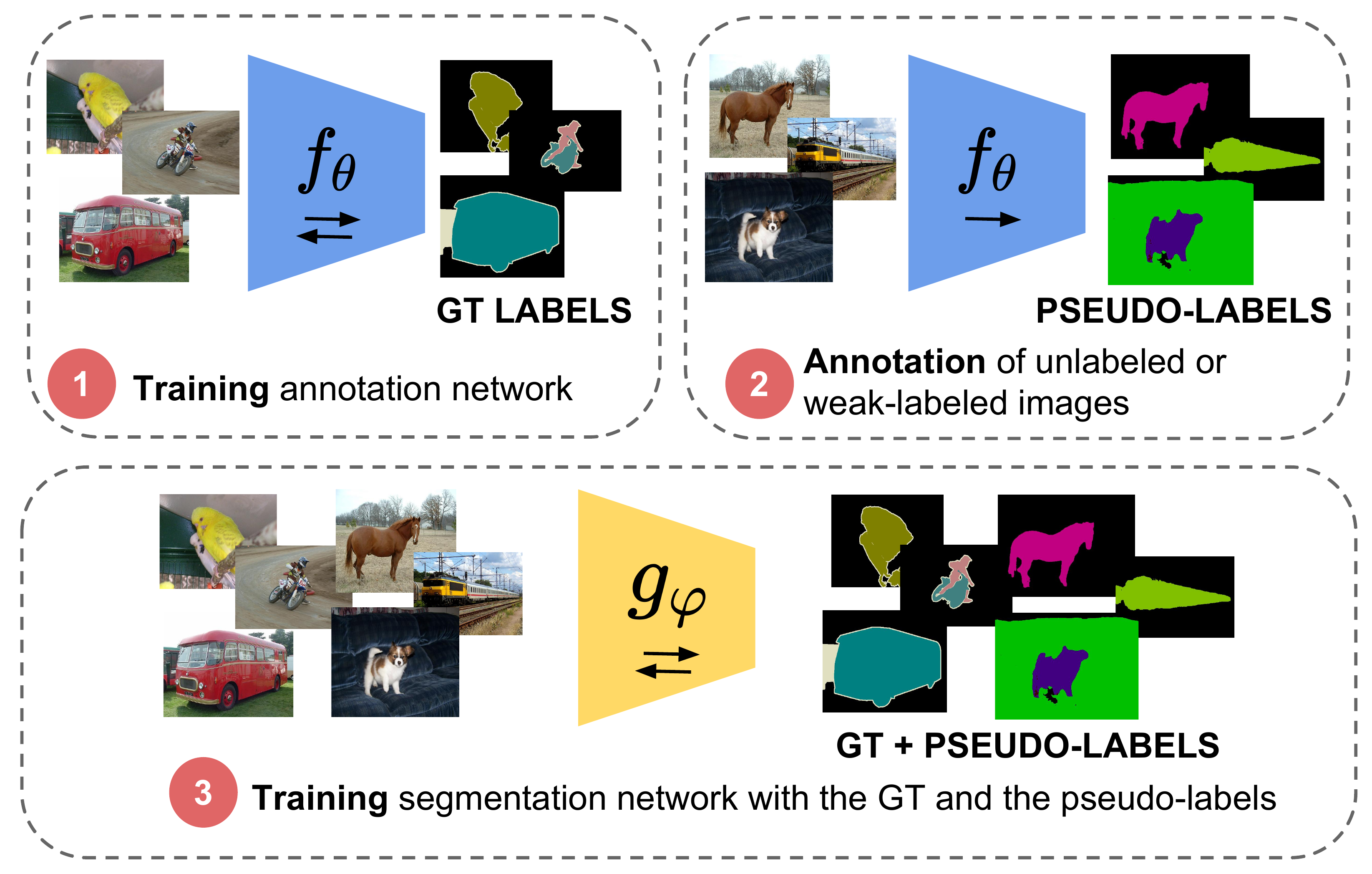}
  \caption{BASIS pipeline from \cite{bellver2019budget} consists of two networks, an annotation network trained with strong supervision, and a segmentation network trained with the union of pseudo-annotations and strong-labeled samples.}

  \label{fig:scheme}
  \vspace{-3.25mm}
\end{figure}

Our sample selection approach is implemented over the semi-supervised scheme we introduced in \cite{bellver2019budget}. \textbf{B}udget-\textbf{A}ware \textbf{S}emi-Supervised Semantic and \textbf{I}nstance \textbf{S}egmentation (BASIS) pipeline consists of two different networks. A first fully supervised model $f_{\theta}$ is trained with strong-labeled samples from the ground truth $(X,Y) = \{(x_{1},y_{1}),...,(x_{N},y_{N})\}$, being $N$ the total number of strong samples. The network $f_{\theta}$ is an annotation network used to predict pseudo-labels $Y' = \{y'_{1}, ...,y'_{M}\}$ for $M$ unlabeled/weakly-labeled samples $X' = \{x'_{1}, ...,x'_{M}\}$. A second segmentation network $g_{\varphi}$ is trained with $(X,Y) \cup (X',Y')$,
as depicted in Figure~\ref{fig:scheme}. 

Our setup consists in working with heterogeneous annotations: strongly-annotated samples~(with pixel-level annotations) and weakly-annotated samples with image-level plus counts~(IL+C). This type of weak annotation consists in indicating the class labels in each image, and the counts of how many times each category appears. 

The architecture for the \textit{segmentation} network $g_{\varphi}$ that we use is the recurrent architecture for instance segmentation RSIS~\cite{salvador2017recurrent}.
We use the open-source code released by the authors. RSIS ~\cite{salvador2017recurrent} consists in an encoder-decoder architecture. 
The encoder is a ResNet-101~\cite{he2016deep}, and the decoder is composed of a set of stacked ConvLSTM's~\cite{xingjian2015convolutional}. At each time step, a binary mask and a class category for each object of the image is predicted by the decoder. The architecture also includes a stop branch that indicates if all objects have been covered. The main feature of this architecture is that its output does not need any post-processing as in object proposal-based methods, where proposals need to be filtered a posteriori. This way, the pseudo-annotations are directly the output of the network itself. 

Regarding the \textit{annotation} network $f_{\theta}$, we modify the RSIS architecture to adapt it to the weakly-supervised setup.
The main difference is that, besides the features extracted by the encoder, the decoder receives at each time step a one-hot encoding of a class category representing each of the annotated instances of the objects in the image. If there are several instances belonging to the same class, a one-hot encoding of that class will be given as input at several time steps, as many as the counts of instances of each depicted class. As we did in our previous work~\cite{bellver2019budget}, we call this architecture \textit{W-RSIS}, where W- refers to the weakly supervised approach.

\section{IoU quality prediction}
\label{sec:iou-quality}

\begin{figure}
  \centering
  \includegraphics[width=\textwidth]{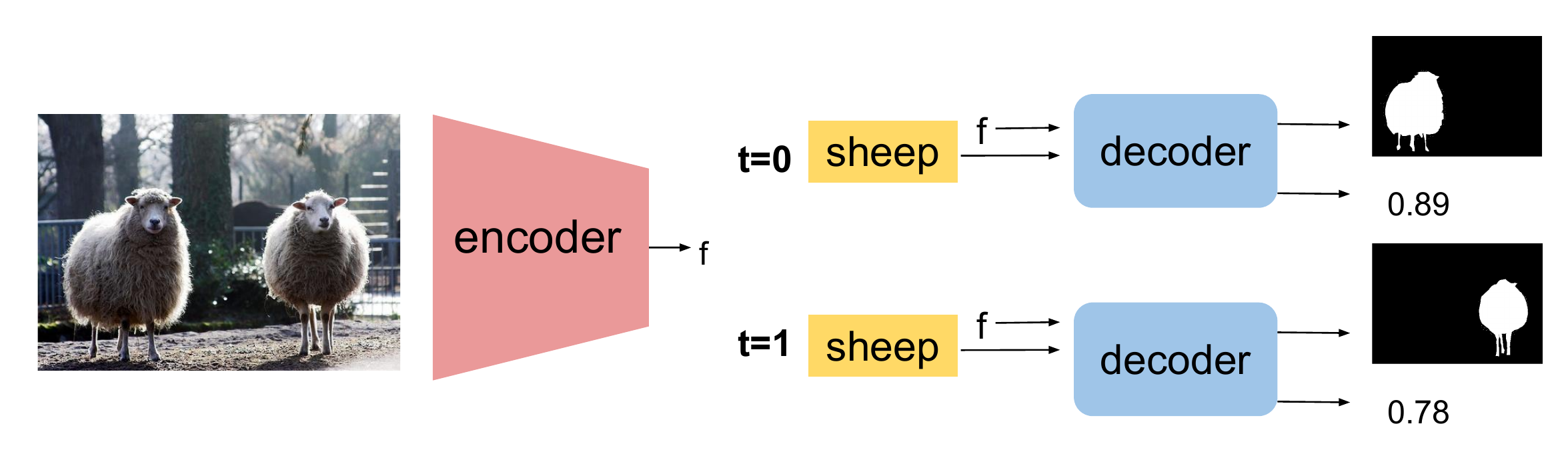}
  \caption{\textit{IoU-W-RSIS} model with the IoU branch.}
  \label{Figure:iou-model-wrsis}
\end{figure}

The main contribution of this work is proposing an additional output to the \textit{W-RSIS} \textit{annotation} network that predicts the quality of each predicted mask. This confidence score can guide an active learning algorithm in choosing which images should be strongly-annotated given a limited budget. We propose to predict the Intersection over Union (IoU) of the predicted masks over a hypothetical ground truth as the guiding signal. As ground truth masks are available for the training data, the model can be trained and the confidence score estimated. The pipeline can be seen in Figure \ref{Figure:iou-model-wrsis}. We call this new architecture \textit{IoU-W-RSIS}. The IoU measures the intersection between two regions divided by its union, and it is a common metric to assess segmentation performance (Equation~\ref{eq:iou}).

\begin{equation}
IoU(A, B)\, =\,\frac{\left |A\cap B  \right |}{\left |A  \right | +  \left |B  \right | - \left |A\cap B  \right |}
\label{eq:iou}
\end{equation}

Therefore, at each time step, our model will segment an object mask of the category fed in the input and predict a confidence score of the segmentation quality. 

\begin{figure}
  \centering
  \includegraphics[width=0.85\textwidth]{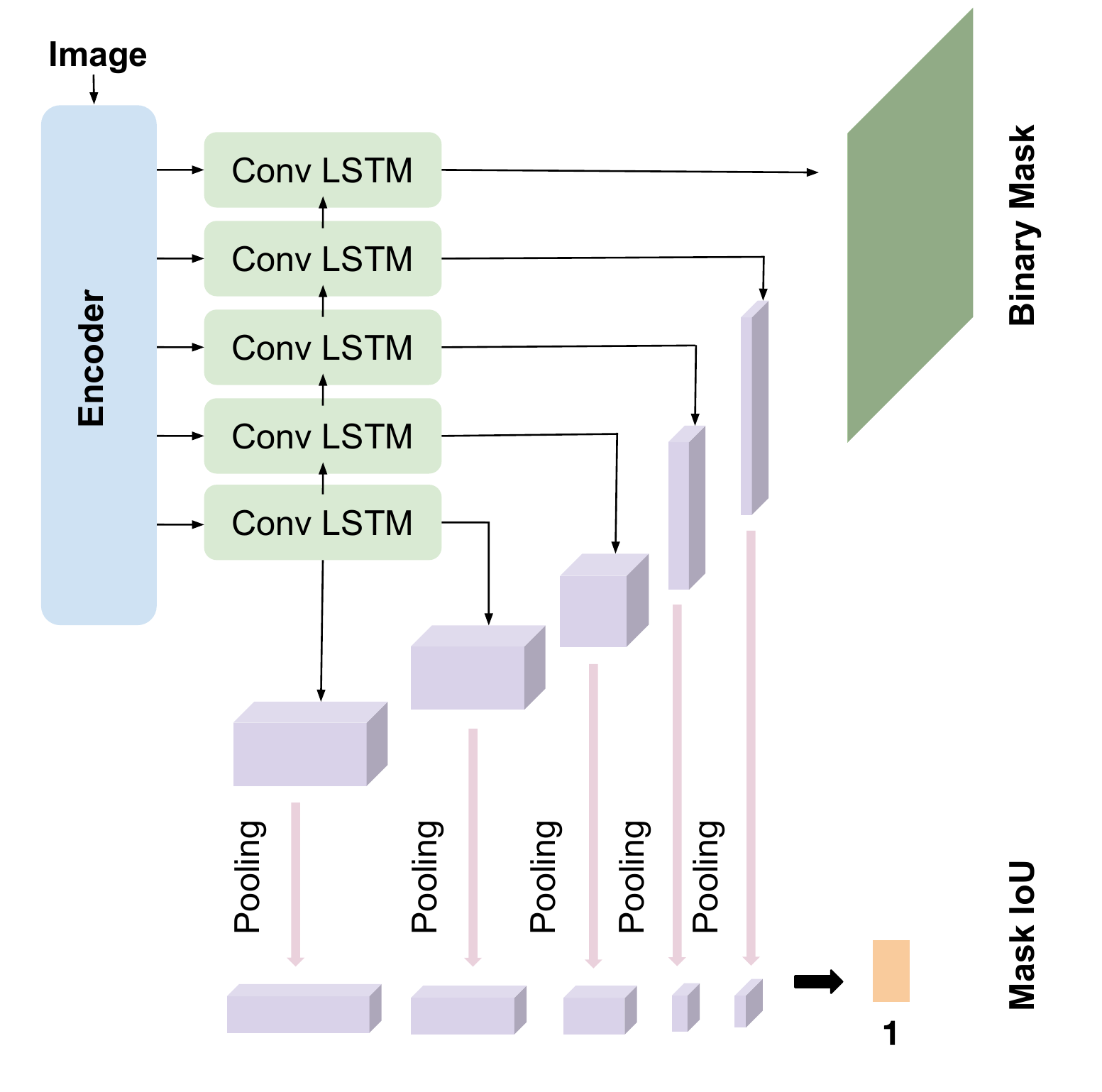}
  \caption{\textit{IoU-W-RSIS} model with the IoU branch for a single time step. The class label is omitted in this figure for clarity.}
  \label{Figure:iou-architecture}
\end{figure}

The architecture that predicts the IoU is depicted in Figure~\ref{Figure:iou-architecture}. A branch for IoU prediction is added to the decoder of the network. This branch aggregates features of a decoder at different spatial resolutions, concatenates them, and computes global average pooling. Afterwards, we add a fully connected layer that predicts the IoU using an L1 regression loss. This loss term is introduced once the segmentation loss has already converged. At that point, the network weights are frozen and only the additional IoU branch is trained for a few epochs. To give more relevance to the predictions of low IoUs, we predict the squared IoU, as suggested in other scenarios in which small values have important relevance, as bounding box offset regression for object detection~\cite{redmon2016you}.

With the proposed architecture, an \textit{IoU Score} for each mask is predicted. In our methodology we use an overall IoU per image instead of individual IoU scores per object. This means that a human annotator will be asked to annotate all object instances from the selected images. Therefore, to compute the \textit{IoU Score} for an image with $M$ objects, we simply average the scores predicted per each object, as seen in Equation~\ref{eq:image_score}.

\begin{equation}
IoU\: Score\, =\, \frac{1}{M}\sum_{i\epsilon M}IoU_{i}
\label{eq:image_score}
\end{equation}

\section{Experiments}
\label{sec:experiments}

The \textit{IoU-W-RSIS} \textit{annotation} network presented in Section~\ref{sec:iou-quality} was tested considering one active learning iteration for the task of instance segmentation. Our experiments aimed at measuring the gain of a IoU guided selection of the images to strongly-annotate, compared with a baseline of random selection as in \cite{bellver2019budget}. We present experiments for the instance segmentation task for the Pascal VOC 2012 benchmark~\cite{everingham2010pascal}.
The standard semi-supervised setup adopted for this benchmark consists in using the Pascal VOC 2012 train images (1464 images) as strong-labeled images, and an additional set (9118 images) from~\cite{hariharan2011semantic} as unlabeled/weak-labeled. 

This section is divided in two subsections, first we focus on the IoU prediction task~(Section~\ref{label:Ablation}), and then we study how to use this score for tackling sample selection~(Section~\ref{label:AL}).

\subsection{IoU Prediction}
\label{label:Ablation}

In this first set of experiments we try several configurations to train the IoU branch of the \textit{IoU-W-RSIS} architecture. We train our proposed annotation network \textit{IoU-W-RSIS} with $N\in\{100,\,200,\,400,\,800,\,1464\}$, where $N$ is the amount of strongly-annotated samples. These $N$ samples are randomly selected from the Pascal VOC 2012 train set~(that has a total of 1464 images). Table~\ref{Table:IoU-MAE} contains the Mean Absolute Error (MAE) computed as the mean of the MAE of \textit{IoU Scores}~(Eq.~\ref{eq:image_score}) of the dataset for the different configurations. The \textit{Baseline} configuration consists in training the IoU branch at the same time as the segmentation branch. In the next row, we freeze the weights of the segmentation network after 150 epochs and only train the IoU branch (until 250 epochs). Finally, we optimize the squared root of the IoU, as small values are specially relevant for this task, and this option leads to the best results. As expected, the MAE tends to decrease from left to right in the table, which corresponds to considering more strongly annotated images.

\begin{table}[]
\centering
\begin{tabular}{@{}lccccl@{}}
\toprule
                              & \textbf{100}    & \textbf{200}         & \textbf{400}    & \textbf{800}     & \textbf{1464} \\       \midrule
Baseline                       & 31.1    &39.8  & 49.3 &47.7  & 51.0          \\
+ Freeze Seg. Network          & 24.8  &\textbf{16.7}  & 19.0  &17.1 & \textbf{16.6}          \\
+ Sqrt Loss                    & \textbf{23.6} &19.5  & \textbf{18.0} & \textbf{17.0}  & \textbf{16.6}               \\ \bottomrule

\end{tabular}
\caption{Mean Absolute Error of IoU prediction.}
\label{Table:IoU-MAE}
\end{table}

\subsection{Mask-guided sample selection}
\label{label:AL}

IoU prediction is used as a criterion to select in which images to invest the annotation budget. These images will then be used to train the \textit{annotation} network for the BASIS pipeline.

The experiment is formulated over a fixed set of 1464 images from Pascal VOC 2012.
Our proposal is first training an \textit{IoU-W-RSIS annotation} network with a few initial random samples~(100), and using that network to pseudo-annotate the remaining samples~(1364). 
Together with the pseudo-annotations, we can predict how confident the network is about the predicted segmentation masks thanks to the IoU branch. From this confidence criterion, we can experiment which samples should be better to fully annotate next. This procedure follows the classic active learning setup, in which the samples to be annotated are iteratively selected. In our case, we experiment with a single iteration, but it could be easily extended to a looped pipeline.

\begin{figure*}
  \centering
  \includegraphics[width=1.0\textwidth]{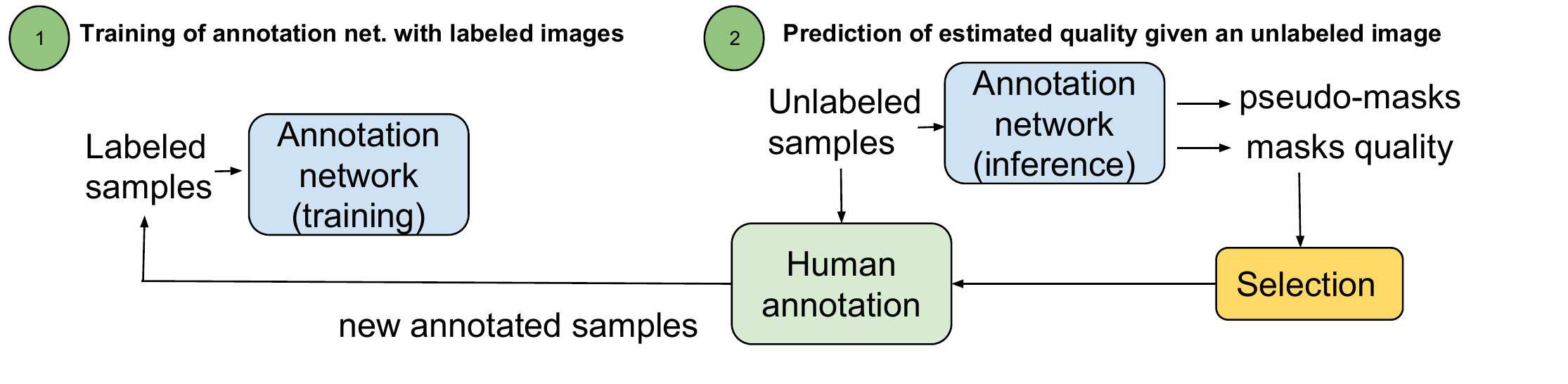}
  \caption{Active Learning pipeline to select next samples to be labeled by a human annotator.}
  \label{Figure:iou-model}
\end{figure*}

\subsubsection{Criterion for sample selection based on IoU:}
\label{sec:criterion}

The experiment in this section explores a criterion for selecting which images should be strongly annotated by a human given the \textit{IoU Scores}. As we want our analysis to focus on the selection criterion only, in this section we will not use the IoU value predicted by our model but the real ground truth value~(\textit{oracle}).

\begin{table}[]
\centering
\begin{tabular}{@{}lccc@{}}
\toprule
                       & \textbf{200}   & \textbf{400}   & \textbf{800}    \\ \midrule
Random subset       & 22.7 $\pm$ 1.8         & 27.1 $\pm$ 0.8                & 34.5 $\pm$ 2.0        \\ \midrule
$\beta=0.0$              & 20.9 $\pm$ 1.5              & 24.1 $\pm$ 0.7           & 29.1 $\pm$ 1.3       \\
$\beta=0.1$              & 22.3 $\pm$ 1.5              & 23.8 $\pm$ 0.6           & 28.6 $\pm$ 0.7         \\
$\beta=0.2$              & 23.3 $\pm$ 0.8              & 24.4 $\pm$ 0.3            & 31.6 $\pm$ 1.1         \\
$\beta=0.3$              & \textbf{23.9 $\pm$ 0.8}     & 26.5 $\pm$ 2.6            & 32.9 $\pm$ 1.4        \\
$\beta=0.4$              & 23.4 $\pm$ 2.7              & \textbf{29.0 $\pm$ 1.3}   & 35.0 $\pm$ 0.6          \\
$\beta=0.5$              & 22.2 $\pm$ 1.1              & 28.9 $\pm$ 0.7            & 35.1 $\pm$ 0.9          \\
$\beta=0.6$              & 22.2 $\pm$ 2.4              & 28.6 $\pm$ 1.3            & \textbf{35.4 $\pm$ 2.4}          \\
$\beta=0.7$              & 22.3 $\pm$ 1.2              & 26.7 $\pm$ 1.3             & \textbf{35.4 $\pm$ 1.4} \\
$\beta=0.8$              & 21.9 $\pm$ 2.0              & 25.3 $\pm$ 1.2            & 33.4 $\pm$ 3.1          \\
$\beta=0.9$              & 20.4 $\pm$ 1.1              & 25.9 $\pm$ 1.1            & 34.8 $\pm$ 1.9          \\
$\beta=1.0$              & 20.3 $\pm$ 1.1              & 25.2 $\pm$ 2.3            & 34.5 $\pm$ 1.3         \\ \bottomrule
\end{tabular}
\caption{Oracle: mean Average Precision~(th=0.5) for different selection criteria~(5 runs for each configuration). }
\label{table:AL_oracle_nn}
\end{table}

Our experiments start with an \textit{IoU-W-RSIS annotation} network trained with only 100 samples, which obtains a performance of 19.0 Average Precision (threshold=0.5). After that, we select another \textit{N'} samples, being $N'\in\{100,\,300,\,700\}$ to make a total of $N\in\{200,\,400,\,800\}$ strongly-annotated samples. The criterion used to select these \textit{N'} samples consists in first defining a set of \textit{IoU Scores} (from 0 to 1.0 in steps of 0.1), that we name $\beta$, and select the \textit{N'} images (being $N'\,\in\{100,\,300,\,700\}$) whose \textit{IoU Scores} are closest to these $\beta$ values. Finally, the samples used to train the \textit{annotation} networks are the 100 initial random images plus these \textit{N'} selected images. The performance obtained with these different subsets is presented in Table~\ref{table:AL_oracle_nn}, which reports the AP (threshold=0.5). All configurations have been trained 5 times, and the reported results are the average with the standard deviation.

The results in Table~\ref{table:AL_oracle_nn} show that there are multiple subsets that outperform a random selection. This means that our selection strategy is effective to reach better performance. We also notice that the optimal predefined \textit{IoU Score} is not fully consistent across different subsets sizes (at $N=800$ the optimal score is 0.6, whereas at $N=200$ the optimal score is 0.3).

\subsubsection{Predicted IoU-selection:}
\label{sec:predicted-iou}

The experiments in Section~\ref{sec:criterion} with the real ground truth IoU~(the \textit{oracle} experiment) showed that choosing samples based on the IoU quality metric leads to better results than performing a random selection. 

In this section, we address the realistic case in which the IoU is predicted by the same annotation network, instead of using the ground truth value as in Section~\ref{sec:criterion}. Table~\ref{table:AL_predicted_nn} shows that for the three set sizes ($N=200,\, 400,\, 800$) better results are also obtained by selecting with the IoU criterion instead of performing a random selection. The optimal \textit{IoU scores} are between 0.3 and 0.6. In fact, we observe a tendency that for smaller subsets, a lower threshold is optimal, whereas for larger subsets, a higher threshold works better. We also observe there is no significant difference between the results obtained with the \textit{oracle} and the predicted IoU configuration.

\begin{table}[]
\centering
\begin{tabular}{@{}lccc@{}}
\toprule
                       & \textbf{200}   & \textbf{400}   & \textbf{800}  \\ \midrule
Random subset       & 22.7 $\pm$ 1.8         & 27.1 $\pm$ 0.8         & 34.5 $\pm$ 2.0        \\ \midrule
$\beta=0.0$              &21.5 $\pm$ 1.1           &23.7 $\pm$ 0.6             &30.1  $\pm$ 1.7        \\
$\beta=0.1$              &21.8 $\pm$ 1.6           &23.7 $\pm$ 0.7             &30.3  $\pm$ 1.7        \\
$\beta=0.2$              &22.6 $\pm$ 0.9           &25.0 $\pm$ 0.8             &29,9  $\pm$ 2.2        \\
$\beta=0.3$              &\textbf{24.0 $\pm$ 1.3}  &26.9 $\pm$ 3.2             &33,5  $\pm$ 3.1       \\
$\beta=0.4$              &23.2 $\pm$ 0.4           &24.8 $\pm$ 2.2             &35.3  $\pm$ 0.9                 \\
$\beta=0.5$              &20.9 $\pm$ 3.1            &25.0  $\pm$ 0.9              &\textbf{37.0 $\pm$ 2.0}      \\
$\beta=0.6$              &20.6 $\pm$ 1.2            &\textbf{27.5 $\pm$ 2.7}     &34.8 $\pm$ 3.0         \\
$\beta=0.7$              &20.3 $\pm$ 1.0            &26.2 $\pm$ 3.1              &36.3 $\pm$ 1.1         \\
$\beta=0.8$              &20.7 $\pm$ 2.1            &26.9 $\pm$ 1.6              &35,9 $\pm$ 2.5         \\
$\beta=0.9$              &20.8 $\pm$ 0.8            &26.1 $\pm$ 1.2              &35,5 $\pm$ 1.1         \\
$\beta=1.0$              &21.1 $\pm$ 1.5            &24.8 $\pm$ 1.5              &34.6 $\pm$ 2.1        \\ \bottomrule
\end{tabular}
\caption{Predicted IoU: mean Average Precision~(th=0.5) for different selection criteria~(5 runs for each configuration). }
\label{table:AL_predicted_nn}
\end{table}

\subsubsection{Sets analysis:} 
\begin{figure*}
  \centering
  \includegraphics[width=0.85\textwidth]{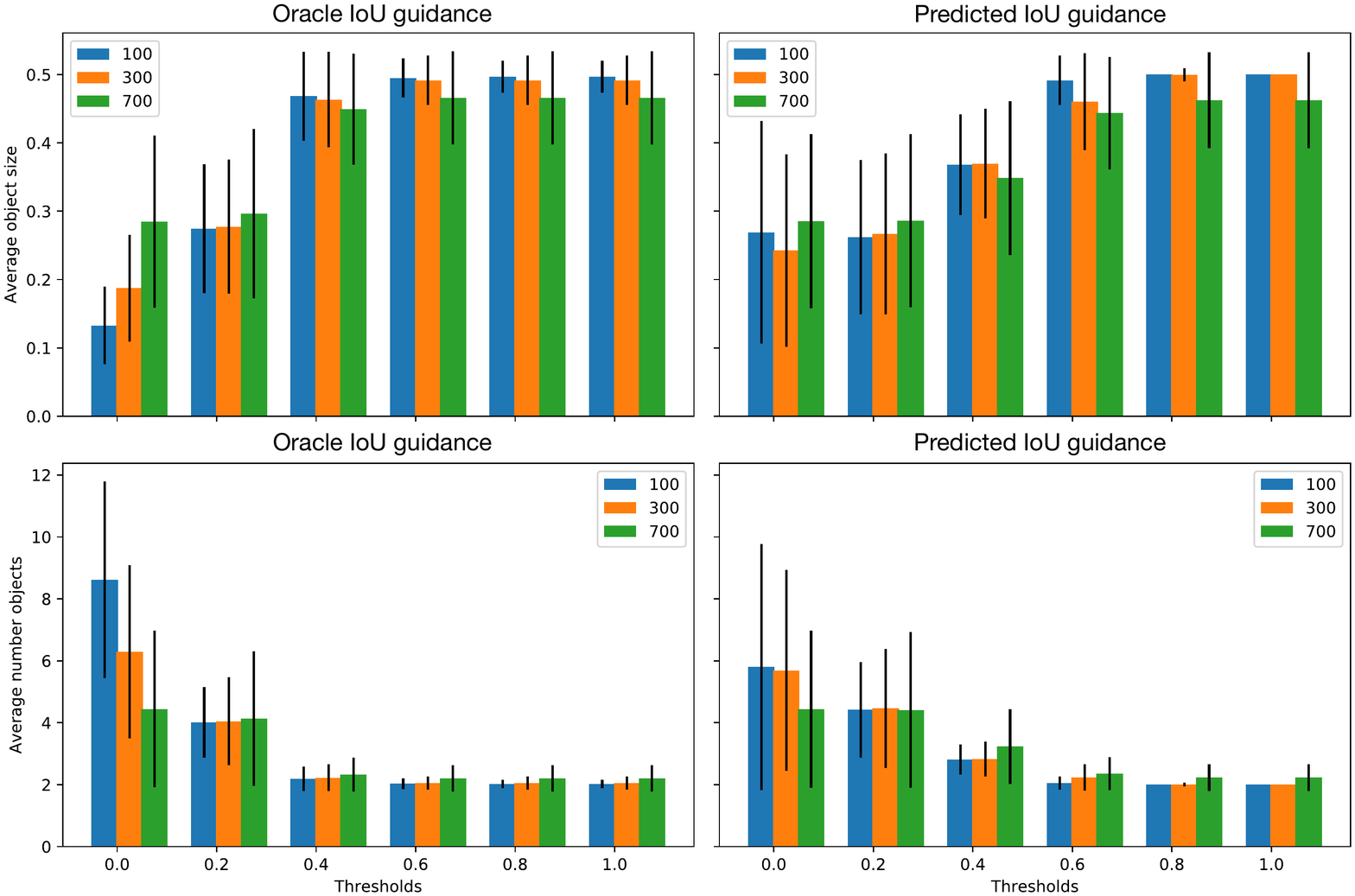}
  \caption{Analysis of the mean object size and number of objects of the selected images.}
  \label{Figure:analysis}
\end{figure*}

In this section we will analyse the properties of the \textit{N'} samples selected based on the sample selection criterion when considering different \textit{IoU Scores} predefined values. We compare the subsets obtained from the \textit{oracle} and the predicted IoU configurations. In Figure~\ref{Figure:analysis} we depict an histogram of the average number of objects per image and the mean size of objects per image for each of the subsets, depending on the predefined \textit{IoU Scores}. The plot has two different columns, the first one belongs to the \textit{oracle} configuration and the second one to the predicted IoU configuration. For both the \textit{oracle} and the predicted IoU configurations, we observe that lower \textit{IoU scores} are related to images with more objects per image and smaller objects. These two scenarios correspond to very challenging cases in object detection, as pointed out by previous works~\cite{girshick2015fast}. Finally, we can observe that the subsets created by the predicted IoU follow a similar distribution to the \textit{oracle} one. 

As we observed in Section~\ref{sec:predicted-iou}, the optimal \textit{IoU Scores} are between 0.3 and 0.6. In Figure~\ref{Figure:analysis} we can see how images associated to these values tend to have a close to the average number of objects per image~(2.8 objects/image). Regarding object size, we observe that objects tend to be neither the largest ones nor the smallest. 

\begin{figure}
  \centering
  \includegraphics[width=0.75\textwidth]{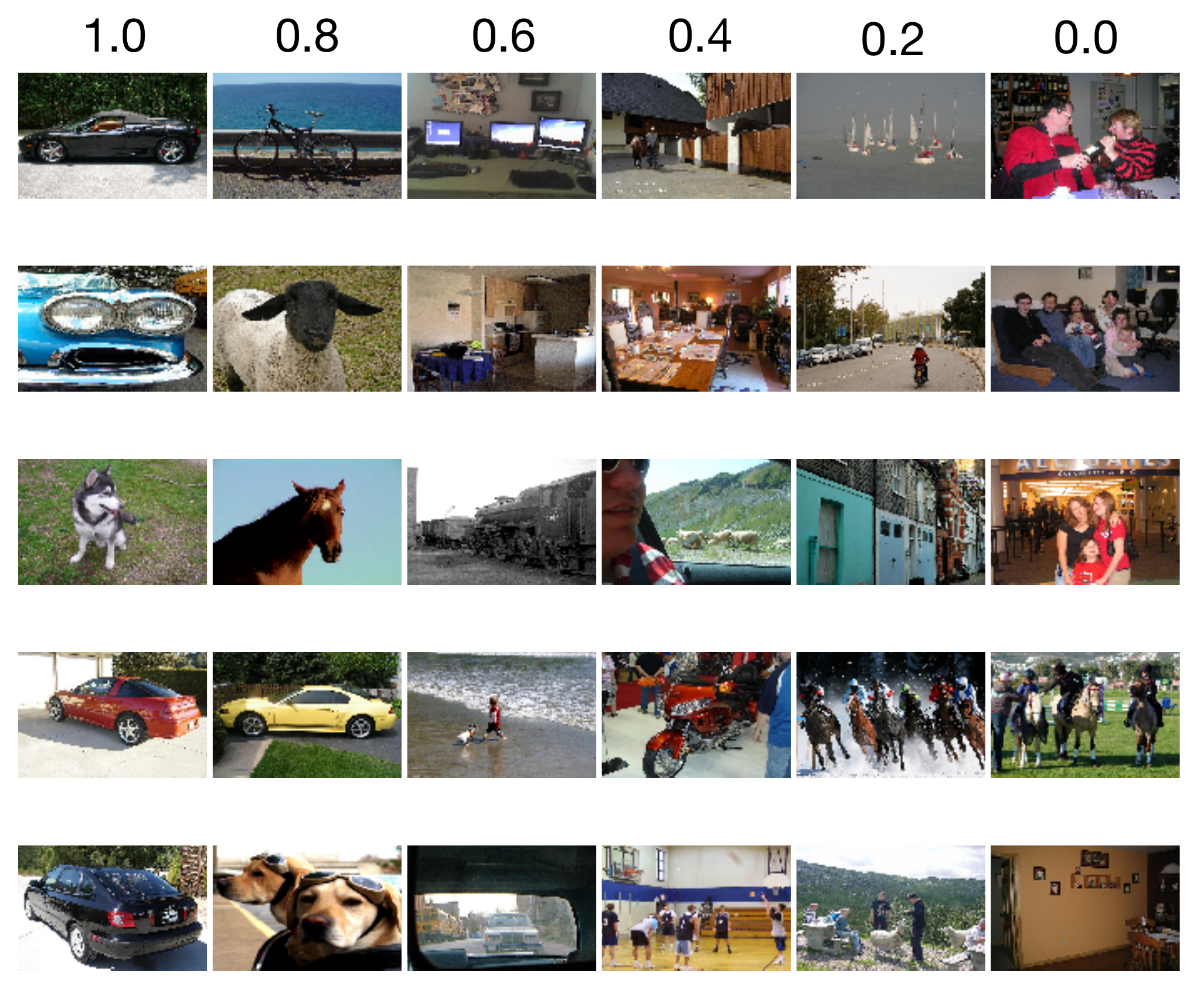}
  \caption{Examples of images of each subset.}
  \label{Figure:examples_vis}
\end{figure}

Figure \ref{Figure:examples_vis} shows some of the selected images when different \textit{IoU Scores} are considered. We observe that at high \textit{IoU Scores} values (0.8 or 1.0), images selected are easy, with only one or two large objects in the image. On the other hand, at low \textit{IoU Scores} (0.0 or 0.2) images have multiple, rather small, instances. As our results indicate, the optimal selected samples to be strongly annotated are those in the middle of the range. These are images that have multiple instances but that are not too complicated to segment. We hypothesize that training with very difficult images can be inefficient if the model is not capable to learn from them, while easy cases do not add much value to the learning process.

\subsubsection{Training of segmentation network:} 

In this section we focus on the final goal of the pipeline, training the \textit{segmentation} network. As a first step, an \textit{annotation} network of $N=200$ is trained with 100 random samples and 100 selected samples (the ones that are closest to the \textit{IoU score} of 0.3, which is the optimal for this set size). The same procedure applies for $N=400$ and $N=800$, with thresholds 0.6 and 0.5 respectively.  Once the \textit{annotation} networks have been trained with the optimal selection of samples given our mask-guided criterion, we use the network to pseudo-annotate the additional Pascal set from \cite{hariharan2011semantic}, a total of 9118 images. Finally, we train the \textit{segmentation} network with the obtained pseudo-annotations and the available strongly-labeled samples.

\begin{figure*}
  \centering
  \includegraphics[width=0.65\textwidth]{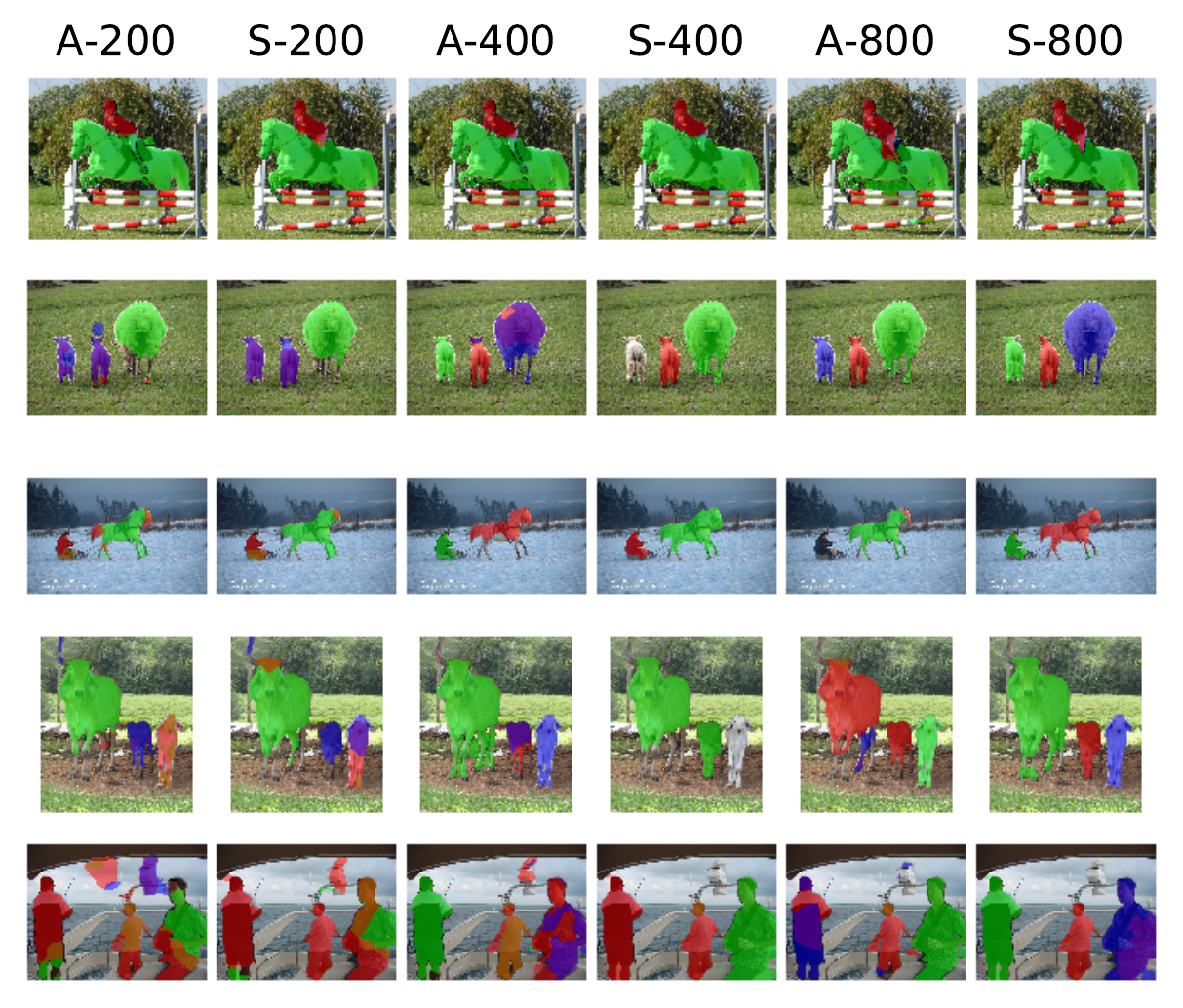}
  \caption{Visualization of Pascal VOC test set for the annotation $f_{\theta_N}$ \textit{(A-)} and segmentation networks $g_{\varphi_N}$ \textit{(S-)}, depending on the number of strong labels used $N\in\{200,\,400,\,800$\}}
  \label{Figure:vis_ann_seg}
\end{figure*}

In Table~\ref{Table:seg-an-comparison} we report the comparison between the random selection of samples and the mask-guided one. We observe that for both the \textit{annotation} and the \textit{segmentation} network performance, the mask-guided selection reaches better results. In Table~\ref{Table:seg-an-comparison-budget} the annotation budgets for each configuration from Table~\ref{Table:seg-an-comparison} are reported. We observe that the mask-guided selection options have a slight higher budget compared to the random ones. This is because the \textit{IoU-W-RSIS} \textit{annotation} network takes as input the image-level labels plus counts. We first predict the \textit{IoU score} for all samples from the Pascal VOC 2012 training set~(1464 samples), and then use those scores to perform the mask-guided selection. Therefore, we need to add the cost for the image-level plus counts labels for \textit{1464-$N$} samples~(as $N$ will be strongly-annotated and already considered in the annotation budget). Figure~\ref{Figure:vis_ann_seg} provides a qualitative comparison between the models obtained from the \textit{annotation} and \textit{segmentation} networks.
We observe that the more strongly-annotated samples($N$), the better quality for the obtained masks. We also observe that the results for the \textit{segmentation} networks are higher than those from the \textit{annotation} networks, proving that the pseudo-annotations are beneficial. 

In Table~\ref{Table:seg-an-comparison-50} we report the mean Average Precision at threshold 0.5 when only the 50\% of the additional set of Pascal VOC \cite{hariharan2011semantic} is weakly-annotated, and therefore the associated budget~(Table~\ref{Table:seg-an-comparison-50-budget}) is lower. In this case we also observe how the configuration with mask-guided selection outperforms the random one. We lead this experimentation to show that at lower annotation budgets, this configuration still works better.

\begin{table}[]
\resizebox{\linewidth}{!}{

\begin{tabular}{lcccccc}
\hline
Selection                      & \multicolumn{3}{c}{Annotation Network}  & \multicolumn{3}{c}{Segmentation Network} \\ \hline
\multicolumn{1}{l|}{}          & 200  & 400  & \multicolumn{1}{c|}{800}  & 200          & 400         & 800         \\ \hline
\multicolumn{1}{l|}{Random (W-RSIS)}    & 22.7 & 27.1 & \multicolumn{1}{c|}{34.5} & 33.3         & 36.8        & 43.8        \\
\multicolumn{1}{l|}{Mask-guided (IoU-W-RSIS)} & 24.0 & 27.5 & \multicolumn{1}{c|}{37.0} & 34.4        & 41.8        & 47.1             \\ \hline
\end{tabular}
}
\caption{Comparison of \textit{annotation} and \textit{segmentation} networks depending on the selection strategy and the number of strongly-annotated samples when all additional set of Pascal is annotated~(9118 images).}
\label{Table:seg-an-comparison}
\end{table}

\begin{table}[]
\resizebox{\linewidth}{!}{

\begin{tabular}{lcccccc}
\hline
Selection                      & \multicolumn{3}{c}{Annotation Network}  & \multicolumn{3}{c}{Segmentation Network} \\ \hline
\multicolumn{1}{l|}{}          & 200  & 400  & \multicolumn{1}{c|}{800}  & 200          & 400         & 800         \\ \hline
\multicolumn{1}{l|}{Random (W-RSIS)}    & 0.55 & 1.11 & \multicolumn{1}{c|}{2.22} & 2.90    & 3.45       & 4.56        \\
\multicolumn{1}{l|}{Mask-guided (IoU-W-RSIS)} & 0.90 & 1.38 & \multicolumn{1}{c|}{2.39} & 3.25    & 3.73       & 4.73            \\ \hline
\end{tabular}
}
\caption{Comparison of \textit{annotation} and \textit{segmentation} networks annotation budget in days when all additional set of Pascal is annotated (9118 images).}
\label{Table:seg-an-comparison-budget}
\end{table}

\begin{table}[]
\resizebox{\linewidth}{!}{

\begin{tabular}{lcccccc}
\hline
Selection                      & \multicolumn{3}{c}{Annotation Network}  & \multicolumn{3}{c}{Segmentation Network} \\ \hline
\multicolumn{1}{l|}{}          & 200  & 400  & \multicolumn{1}{c|}{800}  & 200          & 400         & 800         \\ \hline
\multicolumn{1}{l|}{Random (W-RSIS)}    & 22.7 & 27.1 & \multicolumn{1}{c|}{34.5} & 33.3         & 36.8        & 43.8        \\
\multicolumn{1}{l|}{Mask-guided (IoU-W-RSIS)} & 24.0 & 27.5 & \multicolumn{1}{c|}{37.0} & 34.6              & 38.8            &46.2             \\ \hline
\end{tabular}
}
\caption{Comparison of \textit{annotation} and \textit{segmentation} networks depending on the selection strategy and the number of strongly-annotated samples when 50\% of additional set of Pascal is annotated~(4559 images).}
\label{Table:seg-an-comparison-50}
\end{table}

\begin{table}[]
\resizebox{\linewidth}{!}{

\begin{tabular}{lcccccc}
\hline
Selection                      & \multicolumn{3}{c}{Annotation Network}  & \multicolumn{3}{c}{Segmentation Network} \\ \hline
\multicolumn{1}{l|}{}          & 200  & 400  & \multicolumn{1}{c|}{800}  & 200          & 400         & 800         \\ \hline
\multicolumn{1}{l|}{Random (W-RSIS)}    & 0.55 & 1.11 & \multicolumn{1}{c|}{2.22} & 1.73    & 2.28       & 3.39       \\
\multicolumn{1}{l|}{Mask-guided (IoU-W-RSIS)} & 0.90 & 1.38 & \multicolumn{1}{c|}{2.39} & 2.08    & 2.55       & 3.56            \\ \hline
\end{tabular}
}
\caption{Comparison of \textit{annotation} and \textit{segmentation} networks annotation budget in days when 50\% of additional set of Pascal is annotated~(4559 images).}
\label{Table:seg-an-comparison-50-budget}
\end{table}

\section{Conclusion}
\label{sec:conclusion}

In this work we have extended our previous work on semi-supervised instance segmentation by proposing a novel method to select which samples to strongly-annotate. Our method, based on IoU prediction, outperforms the baseline random selection. We guided a thorough analysis of which samples are best to annotate given the confidence score of the predictions, and we observe that the best samples are those that fall in the mid-range of the IoU scores. With our pipeline, we present a very simple but effective manner to perform sample selection to improve performance at a negligible annotation cost.

\section{Acknowledgments}

This work was partially supported by the Spanish Ministry of Economy and Competitivity under contracts TIN2012-34557 by the BSC-CNS Severo Ochoa program (SEV-2011-00067), and contracts TEC2013-43935-R and TEC2016-75976-R. It has also been supported by grants 2014-SGR-1051 and 2014-SGR-1421 by the Government of Catalonia, and the European Regional Development Fund (ERDF). We would also like to acknowledge the valuable discussions with Victor Campos.





\bibliographystyle{elsarticle-num}
\bibliography{elsarticle-num}








\end{document}